\documentclass[11pt,letterpaper]{article}
\pdfoutput=1

\usepackage{arabtex}
\usepackage{CJKutf8}
\usepackage{utf8}
\usepackage{emnlp2016}
\usepackage{times}
\usepackage[hidelinks]{hyperref}
\usepackage{latexsym}
\usepackage{graphicx}
\usepackage{amsfonts}
\usepackage{footnote}
\usepackage{enumitem}

\emnlpfinalcopy

\makesavenoteenv{tabular}
\makesavenoteenv{table}

\newcommand{\specialcell}[2][c]{%
  \begin{tabular}[#1]{@{}c@{}}#2\end{tabular}}

\title{Sequence-to-sequence neural network models for transliteration}

\author{Mihaela Rosca \\ Google Zurich \\ mihaelacr@google.com
        \And
        Thomas Breuel \\ Google Mountain View, CA \\ tmb@google.com}

\date{}

\begin{document}

\setcode{utf8}

\maketitle

\begin{abstract}
 Transliteration is a key component of machine translation systems and software internationalization. This paper demonstrates that neural sequence-to-sequence models obtain state of the art or close to state of the art results on existing datasets. In an effort to make machine transliteration accessible, we open source a new Arabic to English transliteration dataset and our trained models.
\end{abstract}

\section{Introduction}
Transliteration--the conversion of proper nouns from one orthographic system to another--is an important task in multilingual text processing, useful in applications like online mapping and as a component of machine translation systems. Transliteration is determined by collection of historical accidents, conventions, and statistical regularities: many language pairs have adopted different rules for transliteration over time and many transliterations depends on the origin of a word. These properties make it desirable to look for high-quality, automated machine learning solutions to the problem.

A number of model-based methods for machine transliteration have been developed in the past. Such models assume that character sequences in the source orthography correspond to predictable character sequences in the target orthography, possibly depending on context. Some models additionally assume that such correspondences are influenced by phonetic information. Due to the statistical nature of the problem, components of such models frequently involve parameters and statistical modeling such as hidden Markov models, logistic regression, finite state transducers, and/or conditional random fields (CRFs). Typical examples of such models are \cite{ammar2012transliteration,ganesh2008statistical}; transliteration in such a system is based on an alignment step, followed by a CRF model that performs local string rewriting.

In many areas, including machine translation, end-to-end deep learning models have become a good alternatives to more traditional statistical approaches. This is our motivation for taking a similar approach to transliteration. Unlike statistical models, such end-to-end systems simply take a character string in the source orthography and are trained directly to produce a character string in the target orthography. The closest approaches to the transliteration methods described in this paper are probably found in \cite{rao2015grapheme}, using a bidirectional LSTM models together with input delays for grapheme to phoneme conversion (which can be viewed as a kind of “transliteration” from English to IPA) and \cite{yao2015sequence}, using attentionless sequence-sequence models for the same task.

This paper describes the application of two neural-network based sequence-to-sequence models to transliteration that take principled and general-purpose approaches to alignment and one-to-many or many-to-one correspondences. The first model is based on epsilon insertions and CTC \cite{graves2006connectionist} alignment, the second model is an attentional sequence-to-sequence model commonly used in end-to-end machine translation \cite{bahdanau2014neural}. We report and compare both character (CER) and word error rates (WER) on all problems.
\section{Models and datasets}

\subsection{Epsilon Insertion}
Epsilon insertion (EI) \cite{azawi2013normalizing} is a simple technique for allowing sequence-to-sequence models to produce strings of different lengths from an input string. Epsilon insertion replaces the original problem of transliteration with a similar problem in which the source string has been modified by the insertion of epsilons (which we will represent as `\_`). Transliteration is then performed by an LSTM (possibly bidirectional and deep), and the output is aligned using CTC.
\begin{itemize}[topsep=-0.5pt]
\setlength\itemsep{-0.5em}
\item source string: \begin{CJK}{UTF8}{min}きょうと\end{CJK}
\item source string with epsilons:  \_\_\begin{CJK}{UTF8}{min}き\end{CJK}\_\_\begin{CJK}{UTF8}{min}ょ\end{CJK}\_\_\begin{CJK}{UTF8}{min}う\end{CJK}\_\_\begin{CJK}{UTF8}{min}と\end{CJK}\_\_
\item LSTM output (after training): \_\_ki\_yo\_u\_\_to\_
\item after CTC alignment: \_\_ky\_o\_\_\_\_\_to\_
\end{itemize}

The implementation used for epsilon insertion models is CLSTM\footnote{\url{https://github.com/tmbdev/clstm}}, an open source C++ library. We also open source all our described trained EI.\footnote{\url{https://github.com/googlei18n/transliteration}}.

\subsection{Attentional Sequence-to-Sequence Models}

Attentional sequence-to-sequence models \cite{bahdanau2014neural} (Seq2Seq) work by using an encoder RNN to learn representations of the input sequence and a decoder RNN to produce the output sequence from the hidden representations the encoder created. The attention mechanism allows the decoder to focus on different parts of the input for each time step in the output sequence and can be seen as the analog of the alignment mechanism used in traditional statistical translation models.
Sequence-to-sequence models do not have the implicit monotonicity assumption that unidirectional CTC models do, hence they are more flexible regarding input-output reordering. This is crucial for machine translation, but less important for transliteration where the sound order gets preserved from the source to the target.

For sequence-to-sequence models we experimented with GRU and LSTM cells and assessed the impact of using a bidirectional encoder. As recommended in \cite{sutskever2014sequence}, we feed the input sequence to the encoder in reverse order. For our experiments we used the implementation with an embedding layer provided by TensorFlow \cite{abadi2016tensorflow} .

\subsection{Datasets}
We asses the proposed models on Arabic to English (AR-EN), English to Japanese (EN-JA) transliteration and grapheme to phoneme conversion, specifically English to IPA (EN-IPA). The datasets we used are described in Table \ref{table:datasets}. For Arabic to English transliteration, we introduce a new corpus extracted from Wikipedia: firstly, we created a bilingual dataset of full names from titles of Arabic and English articles referring to the same person; secondly, we used it to learn alignments between name parts to create the final dataset. Since no direction specific information was used in data gathering, the data can be used both for English to Arabic and Arabic to English transliteration. Due to the extraction process, the dataset includes names of various origins (eg. Papadopoulos has Greek origin) and some English tokens contain characters specific to other languages such as: ß, ø, ł.

For the transliteration datasets (EN-JA, AR-EN), the English tokens were lowercased and diacritics removed (è becomes e, ü becomes u). The inputs and outputs of our models are unicode codepoints: the model reads one unicode codepoint at a time in the source string and produces unicode codepoints.

It should be observed that datasets usually used for transliteration differ substantially in their statistical properties from datasets used in other machine learning research. In particular, transliteration datasets usually represent transliterations only once, regardless of how common the word is in the source language. A second issue is that transliteration datasets used for training contain a large number of exceptional words, whose transliteration probably cannot be learned at all. Finally, there are frequently multiple acceptable transliterations for a source word, but these are not usually represented in the training data; that is, many transliterations counted as errors during training and evaluation may be acceptable. In order to remain comparable to prior work in the area, we did not attempt to address these issues in the existing datasets or the datasets we created for this paper; we will return to the question of how this influences performance in the Discussion.
\begin{table}
\small
\centering
\begin{tabular}{ c | c | c | c | c | c }
 Dataset & Size & \specialcell{Avg. \\ input \\ length} & \specialcell{Avg. \\ output \\ length} & \specialcell{Source \\ vocab  \\ size} & \specialcell{Target \\ vocab \\ size} \\
\hline
EN-IPA\footnote{\url{http://www.speech.cs.cmu.edu/cgi-bin/cmudict}} & 123892 &  7.5 & 6.8 &  28 &  38 \\
\hline
EN-JA\footnote{\url{https://github.com/eob/english-japanese-transliteration/}} & 16356 & 10.8 & 6.5 & 29 & 83 \\
\hline
AR-EN\footnote{\url{https://github.com/googlei18n/transliteration}} & 15898 & 6 & 6.8 & 48 & 40
\end{tabular}
 \label{table:datasets}
\caption{Datasets on which we assessed model performance.}
\vspace*{-\baselineskip}
\end{table}

\section{Experimental results}
\subsection{Training and parameters}
For all experiments we used 10\% of data for testing, 10\% of the remaining data for evaluation and the rest for training. All networks were trained using gradient descent with momentum. We used gradient clipping to avoid exploding RNN gradients \cite{pascanu2012difficulty}. EI models use a batch size of 1, gradient clipping norm of 9 and 3 epsilons. When training EI models we randomly varied the learning rate ($10 ^{-5}$ to 0.1), momentum rate (0.5 to 0.99) and number of hidden units (100 to 1000). For sequence-to-sequence models we varied the following hyperparameters: learning rate ($10 ^{-5}$ to 10), momentum rate (0.5 to 0.99), batch size (1 to 50), gradient clipping norm (1 to 10) and number of hidden units (50 to 1000). For both models we trained 1000 networks with different hyperparameter values but a fixed number of layers and chose the one that performed best on the evaluation set and reported performance on the test set. We verified that for each parameter range, optimal performance was reached within the interior of the parameter interval explored.
\subsection{Results}
Our results are described in Tables \ref{table:en2ja}, \ref{table:ar2en} and \ref{table:en2ipa}. Table \ref{table:results} compares our results against models trained on the same datasets. For completeness, we report other transliteration results despite not being directly comparable to our work since they used different datasets. Using statistical phonetic based machine translation \cite{finch2008phrase} reports a 31\% CER for English to Japanese transliteration. \cite{deselaers2009deep} use deep neural networks for Arabic to English transliteration and report a 22.7\% CER, while traditional approaches combined with a single layer perceptron achieved 11.1\% on the same task \cite{freitag2007sequence}.

\begin{table}
\small
\begin{tabular}{ c | c | c | c | c | c }

Model & \specialcell{Nr \\ Layers} &  Bidi & Cell & CER & WER \\
\hline
EI & 1 &  \checkmark &  LSTM &  18.8 & 52.8 \\
\hline
EI & 2 &  \checkmark &  LSTM & \textbf{18.1} & 51.1 \\
\hline
Seq2Seq & 1 &  $\times$  &  GRU &  22.8 & 57.1 \\
\hline
Seq2Seq & 2 &  $\times$ &  GRU  & 20.2 & \textbf{50.2} \\
\hline
Seq2Seq & 3 &  $\times$ &  GRU & 22.2 & 55.4 \\
\hline
Seq2Seq & 1 &  $\times$ &  LSTM & 23.5 & 56 \\
\hline
Seq2Seq & 2 &  $\times$ &  LSTM & 22.5 & 55 \\
\hline
Seq2Seq & 1 &  \checkmark & GRU & 22.6 & 54.6 \\
\hline
Seq2Seq & 2 &  \checkmark & GRU & 20.5 & 51.8
\end{tabular}
\caption{English To Japanese results. Test size: 1780. The RNN size reports the number of units of an individual network. For bidirectional networks the number of units should be multiplied by 2, to account for the network which sees the input in reverse order. Only bidirectional encoders were used for sequence-to-sequence models.}
 \label{table:en2ja}
\vspace{1em}
\begin{tabular}{ c | c | c | c | c | c}

Model & \specialcell{Nr \\ Layers} &  Bidi & Cell & CER & WER \\
\hline
CTC & 1 & \checkmark & LSTM & 22.7 & 79.2 \\
\hline
CTC & 2  & \checkmark & LSTM & \textbf{22.5} & 78.5 \\
\hline
Seq2Seq & 1 & $\times$ &  GRU  & 23.5 & 77.6 \\
\hline
Seq2Seq & 2 & $\times$ & GRU  & 22.4 & \textbf{77.1} \\
\hline
Seq2Seq & 1 & $\times$ & LSTM  & 22.9 & 77.2 \\
\hline
Seq2Seq & 1 & \checkmark & GRU & 22.9 &  78.2 \\

\end{tabular}
\caption{Arabic to English results. Test size: 1590.}
 \label{table:ar2en}

\vspace{1em}

\begin{tabular}{ c | c | c | c | c | c }
Model & \specialcell{Nr \\ Layers} &  Bidi & Cell  & CER & WER \\
\hline
EI & 1 & \checkmark & LSTM & 9.2 & 38.7 \\
\hline
EI & 2 & \checkmark &  LSTM  & 8.1 & 34.2 \\
\hline
Seq2Seq & 1 & $\times$ & GRU  & 7.8 & 28.8 \\
\hline
Seq2Seq & 2 & $\times$ &  GRU  &  \textbf{7.01} &  26.4 \\
\hline
Seq2Seq & 3 & $\times$ &  GRU &  \textbf{7.01} &  26.6 \\
\hline
Seq2Seq & 1 & $\times$ &  LSTM  &  7.40 &  27.0 \\
\hline
Seq2Seq & 2 & $\times$ &  LSTM  &  7.05 &  \textbf{26.2} \\
\hline
Seq2Seq & 1 & \checkmark &  GRU &  7.45 &  27.6 \\
\hline
Seq2Seq & 2 & \checkmark &  GRU &  7.38 &  28.0 \\
\end{tabular}
\caption{English to IPA results. Test size: 12389.}
 \label{table:en2ipa}
\end{table}

\begin{table}
\small
\begin{tabular}{c | c| c| c}
Data & \specialcell{Our\\  WER} &  \specialcell{Their\\ WER} & \specialcell{ Reference}\\
\hline
EN-JA  & \textbf{50.2} & 67.7 &  \cite{benson2009english} \\
EN-IPA & 26.2 & \textbf{21.3} &  \cite{rao2015grapheme}\\
EN-IPA & \textbf{26.2} & 28.6 &  \cite{yao2015sequence} \\
EN-IPA & 26.2 & \textbf{23.5} &  \cite{yao2015sequence}
\end{tabular}
 \label{table:results}
\caption{Comparing results with prior work on the same datasets. All our results reported here use sequence-to-sequence models. For \protect\cite{yao2015sequence} we report results on two models, which we compare to our approach in the Discussion.}
\end{table}

\subsection{Error analysis}
Table \ref{table:errors} shows a list of errors made by a Arabic to English transliteration model. The most common mistake both explored models make on this task is to confuse vowels in the output.  This is expected, given that Arabic has less vowels than English and that often short vowels are not written. Confusing “p” with “b” is another common mistake, accounted by the lack of a corresponding sound for the English “p” in Arabic.
\begin{table}
\small
\centering
\begin{tabular}{ c | c | c }
 Input & Ground truth & Model output \\
 \hline
 \<ينس> & \textbf{j}ens & \textbf{y}ens \\
 \hline
 \<هاوارد> & h\textbf{o}ward & h\textbf{a}ward \\
 \hline
 \<فرج> & far\textbf{a}j & farj \\
 \hline
 \<سميتشك> & sm\textbf{ycze}k & s\textbf{mich}k
\end{tabular}
\caption{Example errors made by an Arabic to English model.}
\vspace{-\baselineskip}
\label{table:errors}
\end{table}

\section{Discussion}
This paper has demonstrated that end-to-end recurrent neural networks achieve high performance on cross script transliteration on common transliteration tasks: EN-JA, EN-IPA and AR-EN.

We have compared epsilon insertion models and attentional sequence-to-sequence models on three benchmarks, and our results show attentional sequence-to-sequence models generally seem to perform better, but not uniformly.
For grapheme to phoneme conversion our attention based sequence-to-sequence models perform better than the attentionless sequence-to-sequence models used in \cite{yao2015sequence}. However, their bidirectional LSTM models which use alignment features outperform our attention based models. The reason can be two fold: the alignment features learned by an alignment specific model help more than the attention implicitly learned by our model, or the simplicity of the LSTM models is advantage against sequence-to-sequence models.

Our results can be extended and potentially improved in a number of ways. A way is by exploring other recurrent network architectures, such as adaptive computation time networks \cite{graves2016adaptive} or classifier combination via boosting or other methods \cite{rao2015grapheme}. Another way is by combining the neural network cost with a target language model cost \cite{chan2015listen}. Since transliteration involves a combination of orthographic and phonetic features, it might be useful to run a separate pronunciation model on the input string and then provide both the grapheme and the phoneme string as input to the transliteration model, mirroring previous non-neural approaches to transliteration \cite{jansche2009named}.

Perhaps one of the most important areas of improvements is that of training data. Right now, transliteration research (including the described work) performs training and evaluation on plain correspondences between strings in two orthographic systems. Such an approach disregards word frequencies, and treats predictions involving alternative, valid transcriptions as errors. Improvements to both the training datasets and the mechanisms for handling multiple predictions will likely result in significant improvements in model performance and correlate more with human evaluations. In addition to improving datasets, our work also points out the need for understanding the relative importance of character and word error rates in evaluating transliterations, since they appear to vary independently.

Given that transliteration is often used as part of machine translation systems, and that such systems themselves are increasingly character based end-to-end system, the question arises whether we need separate transliteration models at all. It appears likely that transliteration will remain a distinct submodule of such systems, since internal graphemic and phonetic representations inside transliteration modules are likely quite different from internal semantic representations required for translation. Experimental evidence from humans also supports the notion of separate and distinct processing of proper nouns and  other nouns \cite{adorni2014electro}.

In addition to demonstrating a simple and novel way of constructing efficient transliteration systems, the benchmarks presented in this paper should be a useful baseline for future work. To this end, we open sourced a new Arabic-English transliteration dataset.

\section{Acknowledgments}
We would like to thank Andy Staudacher, Lara Scheidegger and Vincent Vanhoucke for their support throughout this work.

\bibliography{translit}
\bibliographystyle{emnlp2016}

\end{document}